\documentclass[11pt,a4paper]{article}
\usepackage[hyperref]{acl2020}
\usepackage{times}
\usepackage{latexsym}
\usepackage[shortcuts]{extdash}
\usepackage{url}
\usepackage[T1]{fontenc}
\usepackage{amssymb}
\usepackage{amsmath}
\usepackage{colortbl}
\usepackage{bbm}
\usepackage{breqn}
\usepackage{algorithm}
\usepackage{algpseudocode}
\usepackage{mathptmx}
\usepackage{multicol}
\usepackage{tikz}
\usepackage{booktabs}
\usepackage{tabularx}
\usepackage{dcolumn}
\usepackage{soul}
\newcolumntype{d}[1]{D{.}{.}{#1}}
\usetikzlibrary{shapes.geometric}
\usetikzlibrary{arrows.meta}

\DeclareMathAlphabet{\mathcal}{OMS}{cmsy}{m}{n}
\DeclareSymbolFont{matha}{OML}{txmi}{m}{it}
\DeclareMathSymbol{\varv}{\mathord}{matha}{118}
\newcolumntype{R}{>{\raggedleft\arraybackslash}X}

\usepackage{amsthm}
 
\theoremstyle{definition}
\newtheorem{definition}{Definition}[section]
\newtheorem{proposition}[definition]{Proposition}

\usepackage{microtype}

\aclfinalcopy 
\def\aclpaperid{734} 

\title{Is Your Classifier Actually Biased?\\ Measuring Fairness under Uncertainty with Bernstein Bounds}

\author{Kawin Ethayarajh \\
  Stanford University \\
  \texttt{kawin@stanford.edu} \\}

\date{}

\begin{document}
\maketitle
\begin{abstract}
Most NLP datasets are not annotated with protected attributes such as gender, making it difficult to measure classification bias using standard measures of fairness (e.g., equal opportunity). However, manually annotating a large dataset with a protected attribute is slow and expensive. Instead of annotating all the examples, can we annotate a subset of them and use that sample to estimate the bias? While it is possible to do so, the smaller this annotated sample is, the less certain we are that the estimate is close to the true bias. In this work, we propose using Bernstein bounds to represent this uncertainty about the bias estimate as a confidence interval. We provide empirical evidence that a 95\% confidence interval derived this way consistently bounds the true bias. In quantifying this uncertainty, our method, which we call \emph{Bernstein-bounded unfairness}, helps prevent classifiers from being deemed biased or unbiased when there is insufficient evidence to make either claim. Our findings suggest that the datasets currently used to measure specific biases are too small to conclusively identify bias except in the most egregious cases. For example, consider a co-reference resolution system that is 5\% more accurate on gender-stereotypical sentences -- to claim it is biased with 95\% confidence, we need a bias-specific dataset that is 3.8 times larger than WinoBias, the largest available.
\end{abstract}

\section{Introduction}

NLP models have drawn criticism for capturing common social biases with respect to gender and race \citep{manzini2019black,garg2018word,ethayarajh-2019-rotate}. These biases can be quantified by applying some metric to an embedding space \citep{bolukbasi2016man}, but it is unclear how bias in text embeddings affects decisions made by downstream classifiers. This is because bias is not propagated deterministically: it is possible for minimally biased embeddings to be fed into a classifier that makes maximally biased predictions (and vice-versa). Moreover, recent work has found that WEAT \cite{caliskan2017semantics}, the most popular test of embedding bias, can be easily manipulated to claim that bias is present or absent \citep{ethayarajh2019towards,ethayarajh2019understanding}.

Unlike measuring embedding bias, measuring classification bias is difficult: most NLP datasets are not annotated with protected attributes, precluding the use of standard fairness measures such as equal opportunity \citep{hardt2016equality}. However, manually annotating a large dataset with a protected attribute is slow and expensive. In response to this problem, some have created small datasets annotated with a single protected attribute -- typically gender -- that is used to estimate bias on tasks such as co-reference resolution  \cite{zhao2018gender,kiritchenko2018examining,rudinger2018gender}. This can be done by creating new data or annotating a subset of an existing dataset with the protected attribute. Intuitively, the less data we annotate, the less certain we are that our sample bias is close to the true bias (i.e., what we would get by annotating the entire population).

We propose using Bernstein bounds to express our uncertainty about the sample bias as a confidence interval. First, we show that for standard fairness measures such as equal opportunity and equalized odds \citep{hardt2016equality}, we can define a cost function such that the fairness measure is equal to the difference in expected cost incurred by the protected and unprotected groups. We treat the contribution of each annotated example to the bias as a random variable. Using Bernstein's inequality, we can thus estimate the probability that the true bias is within a constant $t$ of our sample bias. Working backwards, we then derive a confidence interval for the true bias. Treating the ``genres'' of examples in MNLI \citep{N18-1101} as the protected groups and the rate of annotator disagreement as the cost, we offer empirical evidence that our 95\% confidence interval consistently bounds the true bias. 

In quantifying the uncertainty around bias estimates, Bernstein-bounded unfairness helps prevent classifiers from being deemed biased or unbiased when there is insufficient evidence to make either claim. For example, even when the sample bias is positive, it is possible that the true bias between groups is zero. Conversely, a sample bias of zero does not ensure the absence of bias at the population level. Moreover, our findings suggest that most bias-specific datasets in NLP are too small to conclusively identify bias except in the most egregious cases. For example, consider a co-reference resolution system that is 5\% more accurate on gender-stereotypical sentences. For us to claim that this system is gender-biased with 95\% confidence, we would need a bias-specific dataset that is 3.8 times larger than WinoBias \citep{zhao2018gender}, the largest such dataset currently available. Not only does the NLP community need more bias-specific datasets, but it also needs datasets that are much larger than the ones it currently has.

\section{Bernstein-Bounded Unfairness}

In this section, we present the core idea of our paper: Bernstein-bounded unfairness (BBU). In practice, we estimate the bias -- which we call the \emph{groupwise disparity} -- using a small sample of annotated data. Given that this estimate deviates from the true bias (i.e., at the population level), BBU helps us express our uncertainty about the bias estimate using a confidence interval.

\begin{definition}
\label{cost_defn}
Let $c: (y, \hat{y}) \to [0,C]$ denote the cost of predicting $\hat{y}$ when the true label is $y$, where $C \in \mathbb{R}^{+}$ is the maximum cost that can be incurred. 
\end{definition}

\begin{definition}
\label{groupwise}
Let $f: x \to \{-1, 0, +1\}$ denote an annotation function that maps an example to the protected group $A$ ($+1$), the unprotected group $B$ ($-1$), or neither ($0$). The \emph{groupwise disparity} $\delta(f;c)$ between groups $A$ and $B$ is the difference in expected cost incurred by each group: \begin{equation*}
    \delta(f; c) = \mathbb{E}_{a} \left[ c(y_a, \hat{y}_a) \right] - \mathbb{E}_{b} \left[ c(y_b, \hat{y}_b) \right]
\end{equation*}
\end{definition}

\begin{definition}
\label{amortized_defn}
The \emph{amortized disparity} $\hat{\delta}(x_i, f; c)$ for an example $x_i$, given an annotation function $f$ and cost function $c$, is: \begin{equation*}
    \hat{\delta}(x_i, f; c) = \frac{ c(y_i, \hat{y}_i) f(x_i)}{\text{Pr}[f(x) = f(x_i)]}
\end{equation*}
\end{definition}

The amortized disparity of $x_i$ is an estimate of the groupwise disparity based solely on $x_i$. The expectation over all amortized disparities is the groupwise disparity: $\delta(f;c) = \mathbb{E}_x [\hat{\delta}(x, f; c)]$. In practice, given $n$ i.i.d.\ examples $X$, we can take a Monte Carlo estimate of $\delta(f;c)$ by partitioning $X$ into the protected and unprotected groups using $f$ and then calculating the difference in mean cost. An equivalent way of framing this is that we have $n$ random variables $\hat{\delta}(x_1, f; c), ...,  \hat{\delta}(x_n, f; c)$ and we are taking their mean to estimate $\delta(f;c)$. Because examples $X$ are i.i.d., so are the random variables. This means that we can use Bernstein's inequality to calculate the probability that the sample mean $\bar{\delta}$ deviates from the true groupwise disparity $\delta$ by some constant $t > 0$. Where $[-m, m]$ bounds each random variable $ \hat{\delta}(x_i, f; c)$ and $\sigma^2 = \frac{1}{n} \sum \text{Var}[\hat{\delta_i}]$ denotes their variance, by Bernstein's inequality:
\begin{equation}
\label{Bernstein}
\begin{split}
    \text{Pr}[ | \bar{\delta} - \delta |\ > t ] &= \text{Pr}[ | \bar{\delta} - \mathbb{E}[\hat{\delta}] |\ > t ]\\
    &\leq 2 \exp \left( \frac{-nt^2}{2 \sigma^2 + \frac{2}{3} t m} \right) \\
\end{split}
\end{equation}
Since the interval $[-m, m]$ is defined by the frequency of protected and unprotected examples (\ref{amortized_defn}), if we want it to strictly bound the random variable, it should be $[-NC,NC]$, where $N$ is the population size and we assume that there is at least one protected example. However, if this were the interval, (\ref{Bernstein}) could be criticized for being too loose a bound and effectively useless. Therefore we assume that the proportion of the population that is protected and unprotected is bounded and that the lower bounds on these proportions are known.

\begin{definition}
Let $\gamma_A, \gamma_B$ denote the lower bounds of the proportion of the population that is protected and unprotected respectively. Let $\gamma = \min(\gamma_{A}, \gamma_{B})$.
\label{def:gamma}
\end{definition}
Note that the protected group does not necessarily have to be the smaller of the two groups in this setup. We set $\gamma$ to be the lesser of $\gamma_A$ and $ \gamma_B$ to reflect this: if the unprotected group is smaller than the protected group, then $[-m, m]$ will be bounded in $[-C/\gamma_B, C/\gamma_B]$.

\begin{proposition}
Under (\ref{def:gamma}), $[-m, m] \subseteq [-\frac{C}{\gamma}, \frac{C}{\gamma}]$ for any random variable. Using this interval, (\ref{Bernstein}) can be rewritten as:
\begin{equation}
\label{tight_Bernstein}
    \text{Pr}[ | \bar{\delta} - \delta |\ > t ] \leq 2 \exp \left( \frac{-nt^2}{2 \sigma^2 + \frac{2C}{3\gamma} t} \right)
\end{equation}
\end{proposition}

\begin{proposition}
For a given confidence $\rho \in [0,1)$ that the true groupwise disparity $\delta$ falls in the interval $[\bar{\delta} - t, \bar{\delta} + t]$, we can derive $t \in \mathbb{R}^+$ as follows:
\begin{equation}
\begin{split}
    t &= \frac{B + \sqrt{B^2 - 8 n \sigma^2 \log \left[\frac{1}{2} (1 - \rho) \right]}}{2n} \\
    \text{where } B &= -\frac{2 C}{3 \gamma} \log \left[ \frac{1}{2} (1 - \rho) \right]
\end{split}
\label{interval_width}
\end{equation}
This can be derived by rearranging (\ref{tight_Bernstein}) after setting both sides to be equal and then applying the quadratic formula to find the solution to $t$. Note that the width of the confidence interval grows as: (a) the desired confidence $\rho$ increases; (b) the sample size $n$ decreases; (c) $\gamma$ decreases. To our knowledge, Bernstein bounds are the tightest that can be applied here, as they consider the variance of the random variables. We also validated empirically that they are a better candidate than Hoeffding bounds, another common choice.
\end{proposition}

\paragraph{Standard Fairness Measures} How can we use Bernstein-bounded unfairness to derive confidence intervals when the bias metric is demographic parity, equal opportunity, or equalized odds? 
\begin{itemize}
    \item Demographic parity requires that the success rates be equal across all groups. 
    In this case, the cost would be $c(y, \hat{y}) = (1 - \hat{y})$, since the rate of predicting a positive outcome $(\hat{y} = 1)$ must be the same. There are no constraints on the annotation function $f$.
    \item Equal opportunity requires that the true positive rates be equal across groups \citep{hardt2016equality}. The cost would still be $(1 - \hat{y})$ but the annotation function would be $g(x) = f(x) \cdot y(x)$. To use terminology from \citet{hardt2016equality}, including $y(x)$ means that we annotate ``qualified'' examples (i.e., $y(x) = 1$) but not ``unqualified'' ones (i.e., $y(x) = 0$). 
    \item Equalized odds requires that both true and false positive rates be equal across groups \citep{hardt2016equality}. The annotation function would be the same as for equal opportunity but the cost would have to account for differences in false positive rates as well. This could be done by letting $c$ be the zero-one loss. 
\end{itemize}

It is thus possible to define the cost and annotation functions such that the groupwise disparity is equivalent to the bias defined by a common fairness measure. Because of our framing of the problem, we treat the cost as something to be minimized. For example, for equal opportunity, the groupwise disparity was defined as the difference in false negative rates. However, we could set $c(y, \hat{y}) = \hat{y}$ for equal opportunity as well, such that the groupwise disparity is the difference in true positive rates. Both perspectives are equivalent, but one may be more intuitive depending on the use case. 

\section{Proof-of-Concept Experiments}
\label{proofofconcept}
We begin by providing empirical evidence that a 95\% BBU confidence interval consistently bounds the true bias (i.e., population-level groupwise disparity). We conduct our experiments on the MNLI dev set \citep{N18-1101}, used for testing natural language inference. We treat the genres of examples in MNLI as the ``protected groups''. Since the genre annotations are given, we calculate the true bias as the difference in annotator disagreement rates for in-genre versus out-genre examples, effectively treating the human annotators as the classifier whose bias we want to measure. We then use BBU and check whether the true bias falls within the 95\% confidence interval when we estimate the bias using a subset of the data. 

The experiments on MNLI do not measure an important social bias. Rather, they are meant to be a proof-of-concept. We treat the MNLI genres as ``protected groups'' because the protected attribute -- the genre -- is clearly annotated. We use MNLI over smaller datasets annotated with attributes such as gender because this setup -- where the cost is the rate of annotator disagreement -- does not require any model training, making our results easy to replicate. Moreover, this use case illustrates that our conception of bias need not be restricted to social biases -- it can be the difference in cost incurred by any arbitrarily defined groups.

\begin{figure*}
    \centering
    \minipage{0.50\textwidth}
      \includegraphics[width=\linewidth]{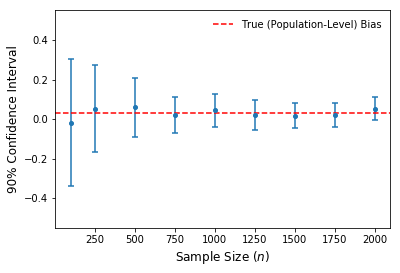}
    \endminipage\hfill
    \minipage{0.50\textwidth}%
      \includegraphics[width=\linewidth]{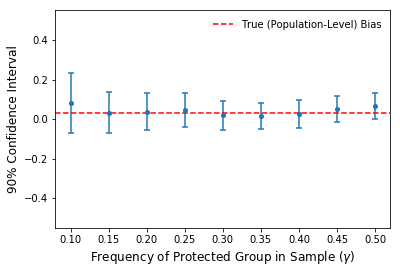}
    \endminipage
    \caption{The true bias (red) for the `government' genre in MNLI and our bias estimates with 95\% confidence intervals (blue), based on a small sample of the data. The bias is defined as the difference in annotator disagreement rates across genres. Our confidence intervals consistently bound the true bias, and the bound grows tighter as the sample size increases (left) and the frequency of the protected group increases (right). On the left, the protected group frequency is fixed at 0.1; on the right, the sample size is fixed at 500.}
    \label{fig:figures}
\end{figure*}

Lastly, we examine how large a bias-specific dataset needs to be in order to conclude that a given classifier is biased. Specifically, we consider a co-reference resolution system that is more accurate on sentences containing stereotypical gender roles. Fixing the confidence level at $\rho = 0.95$, we show that as the magnitude of the sample bias $\bar{\delta}$ decreases, we need a larger bias-specific dataset (i.e., larger $n$) in order to make a bias claim with 95\% confidence.

\subsection{Setup}
\label{ssec:setup}
\begin{table}
\small
\begin{tabular}{lccc}
\toprule
Genre &  In-Genre Cost &  Out-Genre Cost &  $\Delta$ \\
\midrule
facetoface &          0.116 &           0.128 &     $-$0.012 \\
fiction    &          0.122 &           0.128 &     $-$0.006 \\
government &          0.154 &           0.124 &     \ \ \  0.029 \\
letters    &          0.105 &           0.130 &     $-$0.024 \\
nineeleven &          0.115 &           0.129 &     $-$0.014 \\
oup        &          0.132 &           0.127 &     \ \ \  0.005 \\
slate      &          0.147 &           0.125 & \ \ \ 0.022 \\
telephone  &          0.125 &           0.127 &    $-$0.002 \\
travel     &          0.111 &           0.129 &     $-$0.018 \\
verbatim   &          0.146 &           0.125 &     \ \ \ 0.021 \\
\bottomrule
\end{tabular}
\caption{The mean in-genre and out-genre cost for each genre in MNLI, where the cost per example is the rate of annotator disagreement with the gold label.}
\label{tab:descriptive_mnli}
\end{table}

\paragraph{Annotator Disagreement} The MNLI dev set has 10 genres of examples (e.g., `fiction'), with roughly 2000 per genre. Since the genre annotation is known, we treat it as the protected attribute. We define the cost for a given example as the proportion of human annotators whose annotation differs from the gold label. The true bias for each genre (i.e., the groupwise disparity across all data) is the difference in mean cost incurred by the in-genre and out-genre examples. These statistics are in Table \ref{tab:descriptive_mnli}. The annotation function for each genre just samples some in-genre and out-genre examples to be the protected and unprotected groups respectively. In this setup, the ratio of in-genre to out-genre examples is controlled by $\gamma$ (\ref{def:gamma}). We then use this sample to calculate a 95\% confidence interval $[\bar{\delta} - t, \bar{\delta} + t]$. If $\Delta$ in Table \ref{tab:descriptive_mnli} falls within $[\bar{\delta} - t, \bar{\delta} + t]$, then the BBU confidence interval correctly bounds the true bias for that genre.

\paragraph{Gender Bias} For our second experiment, we consider a hypothetical co-reference resolution system $M$ that is more accurate when the input sentence is gender-stereotypical. For example, $M$ might assume that \emph{`doctor'} is always replaced with a male pronoun and \emph{`nurse'} with a female pronoun. The existence of such systems motivated the creation of bias-specific datasets such as WinoBias and WinoGender for co-reference resolution \citep{zhao2018learning,rudinger2018gender}. We define the cost for a given example as the zero-one loss (i.e., $\mathbbm{1}[y \not= \hat{y}]$) so that the true bias corresponds to the difference in accuracy between gender-stereotypical and non-gender-stereotypical sentences. The former is our protected group. Say $\bar{\delta} = 0.05$ -- that is, $M$ is 5 percentage points more accurate on gender-stereotypical sentences. How large must $n$ be for us to claim with 95\% confidence that $M$ is gender-biased (i.e., for $0 \not\in [\bar{\delta} - t, \bar{\delta} + t]$)?

\subsection{Bounding Population-level Bias}

On the MNLI data, even when as few as 100 examples are sampled and used to estimate the bias, a 95\% BBU confidence interval bounds the true bias 100\% of the time. This outcome is the average across all MNLI genres after averaging the results across 20 runs. As seen in Figure \ref{fig:figures}, 95\% BBU bounds also grow tighter as the annotated sample size $n$ increases and the frequency of the protected group $\gamma$ increases from 0.1 to 0.5. Based on the derivation of the interval width in (\ref{interval_width}), both of these trends are expected.

\subsection{Making Claims of Bias}
\label{ssec:making_claims}

In our gender bias experiment, we want to know how large $n$ needs to be such that given $\bar{\delta} = 0.05$, we can say with 95\% confidence that the co-reference resolution system $M$ is gender-biased. In other words, we want to find the smallest $n$ such that $0 \not\in [\bar{\delta} - t, \bar{\delta} + t]$. Since $\bar{\delta} > 0$, we can set $t \gets \bar{\delta}$ and work backwards from (\ref{tight_Bernstein}):
\begin{equation}
    n > \frac{(2\sigma^2 + \frac{2 C}{3 \gamma} \bar{\delta}) \left( -\log \left[\frac{1}{2} (1 - \rho) \right] \right)}{\bar{\delta}^2}
\end{equation}

\begin{figure}
    \centering
    \includegraphics[width=\linewidth]{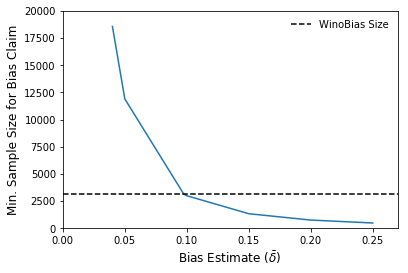}
    \caption{The bias estimate $\bar{\delta}$ of a co-reference resolution system $M$ is calculated on a sample of annotated data. How much data do we need to claim that $M$ is gender-biased with 95\% confidence? The smaller the bias estimate, the more data required. WinoBias, the largest such dataset available, can only be used when $\bar{\delta} \geq 0.0975$.}
    \label{fig:size_needed}
\end{figure}

In our hypothetical scenario, the maximum cost $C = 1$, the bias estimate $\bar{\delta} = 0.05$, and $\rho = 0.95$. We assume that $\gamma = 0.5$, since bias-specific datasets often have equally many protected and unprotected examples. We also assume that the variance is maximal (i.e., $\sigma^2 = (C/\gamma)^2$). 

With these inputs, $n > 11903$: in other words, we would need a bias-specific dataset with at least 11903 examples to claim with 95\% confidence that the system $M$ is biased. This is $\approx 3.8$ times larger than the size of WinoBias \citep{zhao2018gender}, the largest such dataset currently available. In Figure \ref{fig:size_needed}, we plot the amount of data needed against the magnitude of sample bias $\bar{\delta}$. Note that with WinoBias, which has 3160 examples, we could only make a bias claim with 95\% confidence if the bias estimate $\bar{\delta} = 0.0975$ or higher (i.e., if the system $M$ were 9.75 percentage points more accurate on the gender-stereotypical examples in WinoBias).

\subsection{Implications}

It is possible to claim the \emph{existence} of bias in a particular direction without knowing what the true bias is. For example, consider the $\gamma = 0.5$ error bars in Figure \ref{fig:figures} (right): the 95\% confidence interval for the bias faced by the `government' genre in MNLI falls in the range (0.0, 0.12). This means that we are 95\% confident that `government' examples in MNLI face \emph{more} annotator disagreement than other genres, even if we do not know precisely how much more that is. However, as shown in section \ref{ssec:making_claims}, datasets currently used to estimate classification bias in NLP -- such as WinoBias \citep{zhao2018learning} and WinoGender \citep{rudinger2018gender} -- are too small to conclusively identify bias except in the most egregious cases.

There are two possible remedies to this. For one, even though we applied what we thought was the tightest applicable bound, it may be possible to derive a tighter confidence interval for $\delta$. If so, one could use smaller datasets to make bias claims with a high degree of confidence. However, even in this optimistic scenario, current datasets would probably remain insufficient for detecting small magnitudes of bias. The more straightforward remedy would be to create larger bias-specific datasets. Even MNLI, for example, is orders of magnitude larger than WinoBias, suggesting that creating large bias-specific datasets is well within the realm of possibility.

\section{Conclusion}

We first showed that many standard measures of fairness (e.g., equal opportunity) can be expressed as the difference in expected cost incurred by protected and unprotected groups. Given that most bias estimates are made using small samples, we proposed Bernstein-bounded unfairness (BBU) for quantifying the uncertainty about a bias estimate using a confidence interval. Using MNLI, we provided empirical evidence that 95\% BBU confidence intervals consistently bound the true population-level bias. In quantifying this uncertainty, BBU helps prevent classifiers from being deemed biased or unbiased when there is insufficient evidence to make either claim. Although datasets currently used to estimate classification bias (e.g., WinoBias) are undoubtedly a step in the right direction, our findings suggest that they need to be much larger in order to be a useful diagnostic.

\section*{Acknowledgments}

Many thanks to Aidan Perreault, Dallas Card, and Tengyu Ma for providing detailed feedback. We thank Nelson Liu for helpful discussion.

\bibliography{anthology,acl2020}
\bibliographystyle{acl_natbib}

\end{document}